\pdfoutput=1

\PassOptionsToPackage{dvipsnames}{xcolor}

\documentclass[11pt]{article}

\usepackage{authblk}
\usepackage{acl}

\usepackage{times}
\usepackage{latexsym}
\usepackage{array}
\newcolumntype{P}[1]{>{\hspace{0pt}}p{#1}}
\newcolumntype{M}[1]{>{\centering\arraybackslash}m{#1}}
\newcommand{\comment}[1]{}

\usepackage[T1]{fontenc}

\usepackage[utf8]{inputenc}

\usepackage{microtype}
\usepackage{multirow}
\usepackage{enumitem}
\usepackage{graphicx}

\title{Comparison of Lexical Alignment with a Teachable Robot in Human-Robot and Human-Human-Robot Interactions}

\author[1]{Yuya Asano}
\author[1,2,3]{Diane Litman}
\author[2]{Mingzhi Yu}
\author[3]{Nikki Lobczowski}
\author[3,4]{Timothy Nokes-Malach}
\author[1,2]{\\Adriana Kovashka}
\author[1,2,3]{Erin Walker}

\affil[1]{Intelligent Systems Program, University of Pittsburgh, USA}
\affil[2]{Department of Computer Science, University of Pittsburgh, USA}
\affil[3]{Learning Research and Development Center, University of Pittsburgh, USA}
\affil[4]{Department of Psychology, University of Pittsburgh, USA}
\affil[ ]{\texttt{\{yua17, dlitman, miy39, ngl13, nokes, aik85, eawalker\}@pitt.edu}}

\begin{document}
\maketitle
\begin{abstract}
Speakers build rapport in the process of aligning conversational behaviors with each other. Rapport engendered with a teachable agent while instructing domain material has been shown to promote learning. 
Past work on lexical alignment in the field of education suffers from limitations in both the measures used to quantify alignment and the types of interactions in which alignment with agents has been studied.
In this paper, we apply alignment measures based on a data-driven notion of shared expressions (possibly composed of multiple words) and compare alignment in one-on-one human-robot (H-R) interactions with the H-R portions of collaborative human-human-robot (H-H-R) interactions.
We find that students in the H-R setting align with a teachable robot more than in the H-H-R setting and that
the relationship between lexical alignment and rapport is more complex than what is predicted by previous theoretical and empirical work.
\end{abstract}

\section{Introduction and Related Work}

Alignment is the convergence of behavior among speakers and plays an important role in designing the strategies of dialogue systems because it is associated with user engagement \citep{campano2015an} and task success \citep{nenkova-etal-2008-high,callejas2011predicting,lubold2018automated,kory2019exploring}. However, few studies have looked at how this relationship differs in multi-party versus dyadic task-oriented dialogues involving humans and a dialogue agent. This gap prevents us from inferring appropriate alignment strategies for dialogue agents across different group sizes.

Teachable agents act as peers that learners teach via dialogue. These agents have been shown to facilitate learning due to the effect of learning by teaching \citep{leelawong2008designing} and the rapport the agents build with learners \citep{gulz2011extending}. Inspired by theories suggesting that rapport is tied to verbal and non-verbal alignment \citep{lubold2019comfort,tickle1990nature}, prior educational research has explored relationships between rapport with agents and various forms of alignment such as lexical \citep{rosenthal2016robots,lubold2018producing} and acoustic-prosodic \citep{lubold2018producing,kory2019exploring} alignment.

While lexical alignment (the focus of this paper) in educational dialogue has been an active research area, prior studies are limited by 1) alignment measures (repetition of single words \citep{ai2010exploring,friedberg2012lexical,lubold2018producing} or manual annotations of semantics  \citep{rosenthal2016robots}) or 2) dialogue settings (they studied only dyadic interactions with an agent \citep{rosenthal2016robots,lubold2018producing,sinclair2019tutorbot}, dyadic interactions between humans \citep{michel2017measuring,michel_cappellini_2019,MICHEL201950,sinclair2021linguistic}, or multi-party human interactions \citep{friedberg2012lexical}). Multi-party interactions involving an agent remain to be explored with more sophisticated automated measures that can deal with the alignment of a sequence of words.

Therefore, we extend the past work on lexical alignment in educational dialogue in two ways. First, we view lexical alignment as initiation and repetition of shared lexical expressions, which are automatically extracted from dialogue excerpts and can consist of multiple words \citep{dubuisson2021towards}. Along with these metrics, we propose another viewpoint, \textit{activeness}, which quantifies to what extent a speaker is involved in the establishment of shared expressions independent of their partner. Second, we investigate collaborative teaching where two learners co-teach a teachable NAO robot named Emma. We compare how individual learners align with Emma and how alignment relates to rapport with her in this human-human-robot (H-H-R) setting versus in a one-on-one human-robot (H-R) setting. Although, outside of education, some researchers have also investigated H-H-R interactions (e.g., \citealp{kimoto2019lexical}), exploring alignment specifically in educational settings is useful because optimal alignment strategies differ from task to task \citep{dubuisson2021towards}.
Through our comparisons, this paper provides the groundwork for designing different alignment strategies for teachable agents in the H-R and H-H-R settings.

\section{Methodology}

\subsection{Data Collection}\label{data}
\begin{figure}
    \centering
    \includegraphics[width=\linewidth]{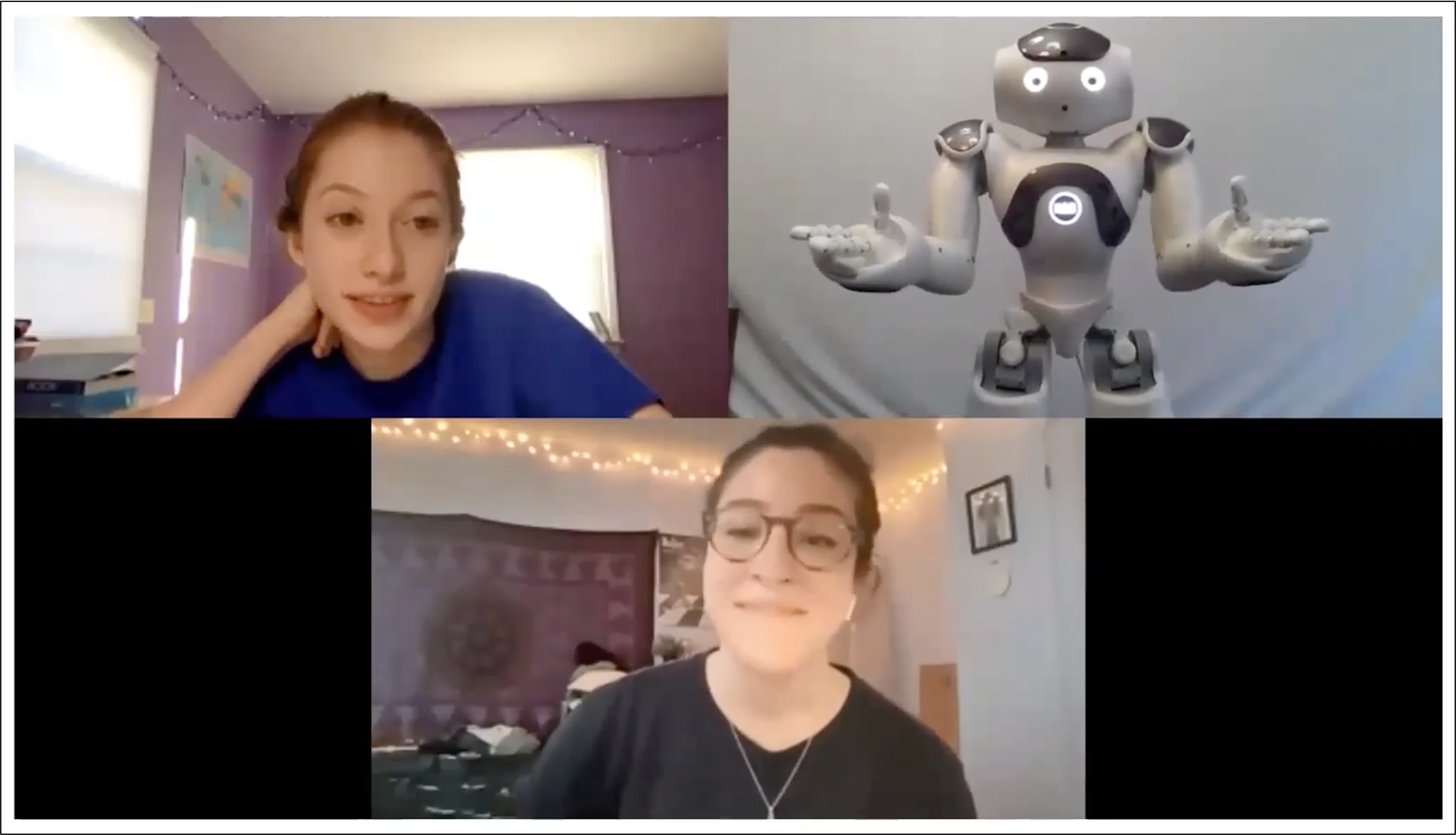}
    \caption{Screenshot of students and Emma in the H-H-R condition.}
    \label{fig:screenshot}
\end{figure}

We recruited 40 undergraduate students from Pittsburgh, USA for an online study (due to COVID) over Zoom. Emma and the student(s) each had their own Zoom window (Figure \ref{fig:screenshot}) and conversed via speech. Students saw ratio word problems on a web application and taught them to Emma for 30 minutes. Each problem consisted of multiple steps, and students had to teach her step-by-step. Emma was designed to guide them by asking a question or making a statement relevant to their response even when they made a mistake. Her responses were pre-authored in Artificial Intelligence Markup Language and were selected based on pattern matching with students' utterances. All students were initially assigned to the H-H-R condition, but they were assigned to the H-R condition if their partner did not show up. We ended up with 12 students in the H-R condition and the remaining 28 in the H-H-R condition to form 14 pairs.
In both conditions, students freely interacted with Emma by pressing and holding a ``push to speak'' button on the application.
In the H-H-R condition, students were also expected to discuss the problems with their partners while teaching Emma, while, in the H-R condition, students had to keep talking to Emma without any discussions with others. An example H-H-R interaction can be found in Appendix \ref{example}. 
We excluded one H-H-R pair from our analysis because one of the students did not talk to either Emma or the partner while working on the problems.

After teaching, learners individually answered survey questions about their perceived rapport with Emma on a six-point Likert scale, ranging from strongly disagree to strongly agree. The survey used four types of rapport measures created by \citet{lubold2018producing}: general rapport measures (three items) based on the sense of connection from \citet{gratch2007creating} and positivity, attention, and coordination rapport measures (four items each, twelve in total) from \citet{sinha2015we} and \citet{tickle1990nature}. The latter twelve items had a higher Cronbach's $\alpha$ (.856) than the general rapport items (.839); thus, we used the average of the positivity, attention, and coordination items to create our rapport metric. The means and standard deviations of our rapport metric were 4.36 and .882 in the H-R condition, and 4.55 and .572 in the H-H-R condition. One-way ANOVA showed no effect of conditions on rapport ($F=.704, p=.407, df=36$).

\subsection{Computing Lexical Alignment}\label{measures}
\comment{
\begin{table}[t]
    \centering
    \begin{tabular}{P{1.47cm}|M{2.66cm}|M{2.66cm}}
    Rapport & H-R (n=12) & H-H-R (n=26) \\
    \hline
    Social & 3.83 (1.06) [.885] & 3.92 (.839) [.826] \\
    Positivity & 4.60 (.985) [.676] & 5.02 (.495) [.640] \\
    Attention & 4.63 (1.04) [.850] & 4.73 (.628) [.379] \\
    Coordination & 3.83 (.943) [.601] & 3.91 (.857) [.801] \\
    \end{tabular}
    \caption{Means (no brackets), SDs (in parentheses), and Cronbach's $\alpha$ (in square brackets) of rapport. One-way ANOVA showed none of four aspects was significantly different across conditions at the $p=.05$ level.}
    \label{rapport}
\end{table}
}

\begin{table}[t]
    \centering
    \begin{tabular}{l|c|c}
    Mean (SD) & H-R (n=12) & H-H-R (n=26) \\
    \hline
    Utterances  & & \\
    \multicolumn{1}{r|}{both speakers}  & 174.8 (27.5) & 59.1 (20.0)\\
    \multicolumn{1}{r|}{student} & 81.3 (13.1) & 27.7 (9.80) \\
    \multicolumn{1}{r|}{Emma}& 93.5 (14.4) & 31.5 (10.4) \\
    \hline
    Tokens & & \\
    \multicolumn{1}{r|}{both speakers} & 2919.0 (466.3) & 1147.0 (388.1) \\
    \multicolumn{1}{r|}{student} & 921.0 (228.7) & 473.0 (227.2) \\
    \multicolumn{1}{r|}{Emma} & 1998.0 (335.9) & 673.0 (210.2) \\
    \end{tabular}
    \caption{Descriptive statistics for H-R dialogues and the two Emma-student portions of H-H-R dialogues.}
    \label{stats}
\end{table}

We manually transcribed all conversations, instead of using Emma's automated speech recognition, because she recorded only while students were holding the ``push to speak'' button. Then, because the measures of lexical alignment below are defined only for dyadic conversations, we manually identified the responder of each utterance in the H-H-R condition (see Appendix \ref{example}) to split each conversation into two Emma-student dialogues and one student-student dialogue. Table \ref{stats} describes the Emma-student dialogue data. 
Although individuals in the H-H-R condition spoke less to Emma than in the H-R condition due to the fixed experiment duration and the dialogue split, this does not affect our measures because they are normalized by the number of shared expressions or tokens. 

The quantification of lexical alignment in the dialogues\footnote{The original dialogues in the H-R condition and the Emma-student dialogue splits in the H-H-R condition.} in this paper relies on a \textit{shared expression}, which is ``a surface text pattern at the utterance level that has been produced by both speakers in a dialogue'' \citep{dubuisson-duplessis-etal-2017-automatic}. A shared expression is initiated by speaker $S$ when used by $S$ first and adapted (thus established as a shared expression) by the dialogue partner later. 
We used the alignment measures derived from shared expressions because mathematical expressions often consist of more than one token, other existing measures compute only repetition, and these measures are shown to be predictive of educational outcomes. Our ratio problems contained fractions and decimals, which cannot be expressed by one word. Indeed, the average lengths of shared expressions were $1.47 \pm .076$ and $1.44 \pm .101$ for the H-R and H-H-R conditions, respectively. Word-based measures such as counting \citep{nenkova-etal-2008-high,friedberg2012lexical,wang-etal-2014-model}, Spearman’s correlation coefficient \citep{huffaker-etal-2006-computational}, regression models \citep{reitter-etal-2006-computational,ward2007automatically}, and vocabulary overlap \citep{campano-etal-2014-comparative} fail to represent the alignment of phrases containing more than one word. Other measures address this issue by leveraging n-grams \citep{michel2017measuring,duran2019align} or cross-recurrence quantification analysis \citep{fusaroli2016investigating} but consider only repetitions in the alignment process as opposed to the measures used in this work \citep{dubuisson-duplessis-etal-2017-automatic,dubuisson2021towards}. Furthermore, \citet{sinclair2021linguistic} have found these measures are correlated with learning and collaboration between human students in collaborative learning.

We employed the set of \textit{speaker-dependent} alignment measures out of the ones proposed by \citet{dubuisson-duplessis-etal-2017-automatic,dubuisson2021towards}\footnote{We used the associated tool available at \url{https://github.com/GuillaumeDD/dialign}.}: Initiated Expression (IE) and Expression Repetition (ER). IE of speaker $S$ (IE\_S) measures orientation (i.e., (a)symmetry) in the alignment process and is defined as $\frac{\textrm{\# expr. initiated by } S}{\textrm{\# expr.}}$. In a dialogue between speakers $S1$ and $S2$, the alignment process is symmetric if $\textrm{IE\_S1} \approx \textrm{IE\_S2} \approx .5$ because $\textrm{IE\_S1} + \textrm{IE\_S2} = 1$. ER of speaker $S$ (ER\_S) captures the strength of repetition and is defined as $\frac{\textrm{\# tokens from } S \textrm{ in new or existing expr.}}{\textrm{\# tokens from }S}$. 

However, IE cannot measure asymmetry or establishment independent of another speaker because, by definition, if IE\_S1 increases, IE\_S2 decreases. This dependence prevents us from observing increased establishment by both speakers. Therefore, we calculated Expression Initiator Difference (IED) \citep{sinclair2021linguistic}, which is given by IED $= |\textrm{IE\_S1} - \textrm{IE\_S2}|$. In addition, we propose a new measure:

\hangindent=1em
\hangafter=1
\noindent \textbf{Expression Establishment} by Speaker $S$ (EE\_S) measures the \textit{activeness} of $S$ in the alignment process in terms of the establishment of new shared expressions. It is given by EE\_S $= \frac{\textrm{\# tokens from } S \textrm{ used to establish new expr.}}{\textrm{\# tokens from }S}$.

In the example dialogue in Appendix \ref{example}, there are ten shared expressions in the Emma-StudentA dialogue split: ``that'', ``can you'', ``can''\footnote{Although the expression ``can'' is part of the longer expression ``can you'', it is counted as a shared expression because it appeared as a free form in Emma's last turn \citep{dubuisson-duplessis-etal-2017-automatic}. I.e., it was not part of ``can you''.}, ``convert'', ``the'', ``days to'', ``days'', ``to'', ``hours'',  and ``hours?''. Of those, Emma started to use three expressions that Student A reused later: ``that'', ``can you'', and ``can''. Thus, IE\_Emma $= \frac{3}{10}$ and IE\_student $= \frac{7}{10}$. These are used to compute IED in the Emma-StudentA dialogue: IED $=|\frac{3}{10} - \frac{7}{10}| = \frac{2}{5}$. ER\_student means the number of tokens in Student A's turns that are taken from Emma's previous turns and therefore parts of shared expressions (these tokens are italicized in Appendix \ref{example}) divided by the total number of tokens Student A spoke to Emma including punctuations. Student A spoke 33 tokens to Emma and devoted four italicized tokens---``can you'' and two ``that''s---to shared expressions. Thus, for Student A, ER\_student $= \frac{4}{33}$. Out of the four, Student A used three tokens to establish new shared expressions ``that'' and ``can you'', so EE\_student $=\frac{3}{33} = \frac{1}{11}$.


\subsection{Alignment Hypotheses}
This study investigates the following hypotheses:

\textbf{H1: Individuals in the H-H-R condition align less with Emma than in the H-R condition.} \citet{Brennan96conceptualpacts} formulated lexical alignment as the establishment of a shared conceptualization, a conceptual pact. In the H-R condition, individuals establish conceptual pacts only with Emma, but, in the H-H-R condition, individuals do so between them through discussion before talking to Emma (see the discussion between students before talking to her in Appendix \ref{example}). This may mean these conceptual pacts are likely to be different from what Emma initially suggested because humans keep updating them, but Emma is not accessible to the updated conceptual pacts (in our case, Emma does not have an ability to intentionally align with humans). Therefore, individuals in the H-H-R condition may tend to use lexicons outside of shared expressions with Emma. 

\textbf{H2: Students feel more rapport with Emma when they align with Emma more (H2-a), she aligns with them more (H2-b), and alignment is more symmetric (H2-c).} Human-human interactions show positive correlations between alignment and rapport \citep{lubold2019comfort,tickle1990nature,sinha2015we}. These are bi-directional; people feel a rapport when aligning with their partners and being aligned by their partners \citep{CHARTRAND2009219}. In human-robot interactions, positive relationships between rapport and non-lexical alignments such as acoustic-prosodic \citep{lubold2018producing,kory2019exploring} and movement \citep{choi2017movement} have also been found. We thus expect lexical alignment positively correlates with rapport in both conditions. We also anticipate a symmetric alignment process positively correlates with rapport because human-human interactions are more symmetric than human-agent ones \citep{dubuisson2021towards} and past work increased rapport by imitating human alignment behavior.

\textbf{H3: Lexical alignment is more strongly correlated with rapport with Emma in the H-R condition than in the H-H-R condition.} As shown in \citet{yu2019investigating}, \citet{levitan-etal-2012-acoustic}, and \citet{namy2002gender}, the alignment process in H-H-R dialogues may also depend on other factors including the gender diversity of the party. Thus, in the H-H-R condition, lexical alignment alone may not be as predictive of rapport as in the H-R condition.

\section{Results and Discussion}

\begin{table}[t]
    \centering
    \begin{tabular}{c|c|c}
    \centering Mean (SD) & H-R (n=12) & H-H-R (n=26) \\
    \hline
    ER\_student** & .594 (.052)  & .462 (.077) \\
    EE\_student & .189 (.033) & .171 (.050) \\
    ER\_Emma** & .494 (.031) & .421 (.079) \\
    EE\_Emma* & .087 (.024) & .119 (.037) \\
    IED & .155 (.111) & .158 (.127) \\
    \end{tabular}
    \caption{Descriptive statistics of lexical alignment measures. Measures marked with * and ** are significantly different across conditions at $p<.05$ and $p<.01$ (two-tailed), respectively.}
    \label{alignment_mean}
\end{table}

\paragraph{Individual alignment across  H-R and H-H-R conditions (H1).}
We tested H1 by comparing means of ER\_student and EE\_student across conditions with one-way ANOVA. Table \ref{alignment_mean} partly supports H1. Individuals in the H-R condition repeated shared expressions (i.e., higher ER\_student) more than in the H-H-R condition, but they were equally likely to establish shared expressions (i.e., no difference in EE\_student) across conditions. 

\comment{
\begin{table*}[t]
    \centering
    \begin{tabular}{c|c|c|c|c|c|c}
    \hline
    \multicolumn{2}{c|}{Correlation (p-value)} & ER\_student & EE\_student & ER\_Emma & EE\_Emma & IED \\
    \hline
    & Social & .098 (.761) & -.074 (.820) & -.053 (.871) & -.039 (.905) & -.144 (.655) \\
    H-R & Positivity & -.374 (.232) & -.158 (.624) & .180 (.576) & .388 (.213) & -.384 (.217) \\
    (n=12) & Attention & .099 (.760) & -.053 (.870) & .011 (.974) & -.053 (.870) & -.476 (.118) \\
    & Coordination & .109 (.735) & -.205 (.524) & -.208 (.516) & -.113 (.727) & -.370 (.236) \\
    \hline \hline
    & Social & -.459* (.018) & -.420* (.033) & .129 (.531) & .168 (.412) & -.456* (.019) \\
    H-H-R & Positivity & -.529** (.005) & -.377 (.058) & .261 (.199) & .257 (.205) & -.297 (.141) \\
    (n=26) & Attention & -.318 (.113) & -.436* (.026) & .340 (.089) & .500** (.009) & -.353 (.077) \\
    & Coordination & -.447* (.022) & -.264 (.193) & .377 (.058) & .503** (.009) & -.405* (.040) \\
    \hline
    \end{tabular}
    \caption{Spearman's correlations between alignment measures and rapport in the H-R (top) and H-H-R (bottom) conditions. Correlations marked with * and ** are significant at $p<.05$ and $p<.01$ (2-tailed), respectively.}
    \label{correlations}
\end{table*}
}

\begin{table*}[t]
    \centering
    \begin{tabular}{c|c|c|c|c|c}
    Estimate of $\beta_3$ (p-value) & ER\_student & EE\_student & ER\_Emma & EE\_Emma & IED \\
    \hline
    Rapport & -0.54 (.901) & 9.09 (.171) & 3.10 (.652) & -0.83 (.928) & 3.72 (.057) \\
    \end{tabular}
    \caption{Coefficients of interaction terms ($\beta_3$).}
    \label{regression}
\end{table*}

\comment{
\begin{table*}[t]
    \centering
    \begin{tabular}{c|c|c|c|c|c}
    Pearson's r (p-value) & ER\_student & EE\_student & ER\_Emma & EE\_Emma & IED \\
    \hline
    Social & -.275 (.094) & -.352* (.030) & .109 (.513) & .088 (.597) & -.500** (.001) \\
    Positivity & -.461** (.004) & -.351* (.031) & .004 (.982) & .280 (.088) & -.366* (.024) \\
    Attention & -.231 (.164) & -.352* (.030) & .192 (.248) & .284 (.084) & -.520** (.000) \\
    Coordination & -.284 (.084) & -.274 (.091) & .230 (.164) & .294 (.073) & -.463** (.003) \\
    \end{tabular}
    \caption{Pearson's correlations between alignment measures and rapport. Correlations marked with * and ** are significant at $p<.05$ and $p<.01$ (2-tailed), respectively.}
    \label{correlations}
\end{table*}
}

\begin{table*}[t]
    \centering
    \begin{tabular}{c|c|c|c|c|c}
    Pearson's r (p-value) & ER\_student & EE\_student & ER\_Emma & EE\_Emma & IED \\
    \hline
    Rapport & -.315 (.054) & -.331* (.043) & .214 (.198) & .343* (.035) & -.573** (.000) \\
    \end{tabular}
    \caption{Pearson's correlations between alignment measures and rapport. Correlations marked with * and ** are significant at $p<.05$ and $p<.01$ (2-tailed), respectively.}
    \label{correlations}
\end{table*}

\begin{table}[t]
    \centering
    \begin{tabular}{c|c|c}
    \centering Pearson's r & H-R (n=12) & H-H-R (n=26) \\
    \hline
    ER\_student & -.145 & -.406 \\
    EE\_student & -.457 & -.285 \\
    ER\_Emma & .008 & .461 \\
    EE\_Emma & .195 & .407 \\
    IED & -.723 & -.529 \\
    \end{tabular}
    \caption{Comparison of Pearson's correlations between alignment measures and rapport across conditions.}
    \label{cond_corr}
\end{table}

\comment{
\begin{table*}[t]
    \centering
    \begin{tabular}{c|c|c|c|c}
    ER\_student & EE\_student & ER\_Emma & EE\_Emma & IED \\
    \hline
    -.145, -.504 (.299) & -.457, -.369 (.788) & -.002, .404 (.274) & .190, .388 (.581) & -.723, -.434 (.253) \\
    \end{tabular}
    \caption{Comparison of Pearson's correlations between alignment measures and rapport in H-R (left) and H-H-R (right) conditions and p-values (in parentheses).}
    \label{cond_corr}
\end{table*}
}

\paragraph{Correlations of alignment with rapport across conditions (H2 and H3).}
To test H2 and H3, first, we fit the regression equation with an interaction between the conditions and an alignment measure:
$
R = \beta_0 + \beta_1 * HHR + \beta_2 * A + \beta_3 * HHR * A
$
where R is the rapport measure, A is an alignment measure, and HHR is 1 for students in the H-H-R condition; otherwise 0. Table \ref{regression} shows that $\beta_3$ is not significant for none of the alignment measures, meaning that the correlations between rapport and alignment are in the same direction regardless of the conditions. 

Therefore, we used all data to compute Pearson's correlations between rapport and alignment (see Table \ref{correlations}). The significant negative correlation between rapport and IED supports H2-c. H2-b is not fully supported because, although EE\_Emma is correlated positively with rapport, ER\_Emma is not. In addition, surprisingly, we found evidence for the opposite of H2-a; EE\_student has a negative correlation with rapport. Further analysis revealed IE\_Emma is significantly negatively correlated with rapport ($r=-.490, p=.002$). This means students felt less rapport when they established more shared expressions relative to Emma and aligns with the findings on EE. 

Finally, we compared Pearson's r between lexical alignment and rapport in the two conditions using Fisher transformation \citep{snedecor1980statistical} to test H3. It was not validated because there was no significant difference between the two conditions in Table \ref{cond_corr}.


These results may be because perceived success in communication with Emma characterized by her accidental alignment leads to high rapport and low alignment by students. As \citet{BRANIGAN20102355} and \citet{dubuisson-duplessis-etal-2017-automatic} reported, students might have (either consciously or unconsciously) expected they should establish shared expressions more than Emma due to her limited linguistic capacity. Thus, they might have started with an asymmetric alignment process. When Emma was stuck, they might have kept this strategy because they thought she did not understand them, resulting in decreased rapport. In contrast, as Emma established new shared expressions by accident, students might have thought she was following new information like humans, that she cared what they said, and that they were in sync, leading to more positivity, attention, and coordination rapport, respectively \cite{tickle1990nature}. They may have also changed their alignment strategy to a more symmetric one (i.e., decreased alignment by students) that they usually use while interacting with humans. 

\subsection{Limitations}
This study has several limitations. First, the limited number of participants (38 in total) might limit the detection of all correlations. 
Moreover, the comparison between the H-R and H-H-R conditions has low statistical power because the H-R condition had fewer than half of the participants in the H-H-R condition. It might have been biased because the assignment to the conditions was not fully random as well. Next, alignment measures may need contextual adjustments. For example, one math problem included both ``three hours'' and ``three-fortieths of battery''. Although ``three'' in these numbers refers to different entities, our measures saw it as a shared expression. Finally, some lexicons came from the problem prompt rather than the group conversation. 

\section{Conclusion}

We examined relationships between lexical alignment and rapport with a teachable agent in one-on-one (H-R) and collaborative (H-H-R) teaching. Our methods expand prior literature by comparing alignment behavior in H-R and H-H-R settings and extending recent work by \citet{dubuisson2021towards} 
to the speaker-level act of \textit{activeness} 
in the alignment process. Our results imply learners' lexical alignment with teachable agents may not always increase rapport with a teachable agent, unlike predictions from alignment theories \citep{lubold2019comfort,tickle1990nature} largely based on human-human interactions. Future work can expand our work by looking at the role of H-H portions of H-H-R interactions in their H-R portion and the effect of miscommunication as an intermediate variable on the negative correlations between rapport and learners' alignment and by extending the measures to multi-party settings without disentanglement.

\section*{Acknowledgements}
We would like to thank anonymous reviewers for their thoughtful comments on this paper. This work was supported by Grant No. 2024645 from the National Science Foundation, Grant No. 220020483 from the James S. McDonnell Foundation, and a University of Pittsburgh Learning Research and Development Center internal award.

\bibliography{anthology,custom}
\bibliographystyle{acl_natbib}

\appendix

\section{Sample Dialogue and Lexical Alignment}
\label{example}

\begin{table*}[t]
    \centering
    \begin{tabular}{c|p{11cm}|c}
        \hline
        Speaker & Utterance & Responder \\
        \hline
        \textbf{Emma:} & \textbf{Now \textcolor{red}{that} I know how long one battery will last, \textcolor{blue}{can you} help me figure out how many batteries I need total?} & \textbf{Student A}\\
        Student B: & Oh, okay. & Student A\\
        Student A: & Okay. Next ... & Student B\\
        Student B: & Okay. & Student A\\
        Student A: & So one whole battery lasts three and three quarters of an hour. & Student B \\
        Student B: & Three and three quarters of an hour. Oh my gosh. Emma, you're making this difficult on us. & Student A\\
        Student A: & I think we need her... Oh man.  & Student B \\
        Student B & Because it's in days now. So she has to figure out...  & Student A \\
        Student A: & You have to do like dimensional analysis. & Student B \\
        Student B & I think she... Yeah. She has to convert days to hours. I think that might be the easiest thing for her. & Student A \\
        Student A: & Okay. & Student B \\
        Student B: & So she has to divide...  & Student A \\
        Student A: &Wait, is she going to remember that?  & Student B \\
        Student B: & Oh, I don't know.  & Student A \\
        Student A: & Okay. I'm going to ask if she knows how to convert days to hours. & Student B \\
        Student B: & Okay. & Student A \\
        \textbf{Student A:} & \textbf{Okay, Emma. \textcolor{blue}{\textit{Can you}} \textcolor{YellowOrange}{convert} \textcolor{ForestGreen}{the} number of \textcolor{Gray}{days to} \textcolor{ForestGreen}{the} number of \textcolor{Magenta}{hours?}} & \textbf{Emma} \\
        \textbf{Emma:} & \textbf{So I know how long I'll be gone in \textcolor{Gray}{\textit{days}}, but how long \textcolor{ForestGreen}{\textit{the}} battery lasts is in \textcolor{Magenta}{\textit{hours}}. So first I should change \textcolor{ForestGreen}{\textit{the}} \textcolor{Gray}{\textit{days to}} \textcolor{Magenta}{\textit{hours?}}} & \textbf{Student A} \\
        \textbf{Student A:} & \textbf{Yes, Emma. \textcolor{red}{\textit{That}}'s correct.} & \textbf{Emma} \\
        \textbf{Emma:} & \textbf{So I \textcolor{blue}{\textit{can}} \textcolor{YellowOrange}{\textit{convert}} two and three quarters \textcolor{Gray}{\textit{to}} an improper fraction, eleven over four. And then I \textcolor{blue}{\textit{can}} multiply it by twenty four \textcolor{Magenta}{\textit{hours?}}} & \textbf{Student A} \\
        Student B: & She did all the work. & Student A \\
        \textbf{Student A:} & \textbf{Yes, Emma. \textcolor{red}{\textit{That}}'s correct.} & \textbf{Emma} \\
        \hline
    \end{tabular}
    \caption{Lexical alignment in the Emma-studentA portion of the dialogue (bolded utterances). A responder is a speaker who responded to the utterance. Speakers initiated colored but not italicized expressions and repeated the italicized ones. Contractions were tokenized as two tokens (e.g., That's to ``That'' and ``'s''). Punctuation was treated as one token but did not constitute a shared expression by itself.}
    \label{example_dialogue}
\end{table*}

\end{document}